\documentclass[letterpaper]{article} % DO NOT CHANGE THIS
\usepackage[authoryear]{natbib}  % 启用作者-年份格式
\usepackage{aaai2026}  % DO NOT CHANGE THIS
\usepackage{times}  % DO NOT CHANGE THIS
\usepackage{helvet}  % DO NOT CHANGE THIS
\usepackage{courier}  % DO NOT CHANGE THIS
\usepackage[hyphens]{url}  % DO NOT CHANGE THIS
\usepackage{graphicx} % DO NOT CHANGE THIS
\def\UrlFont{\rm}  % DO NOT CHANGE THIS
\usepackage{caption} % DO NOT CHANGE THIS AND DO NOT ADD ANY OPTIONS TO IT
\usepackage{booktabs}
\usepackage{xcolor}
\usepackage{multirow}
\usepackage{amsmath}
\setcitestyle{round} % 圆括号 (默认)
\setcitestyle{aysep={,}} % 作者-年份分隔符：(Smith, 2020)
\setcitestyle{yysep={;}} % 多文献分隔符：(Smith, 2020; Jones, 2021)

\setcitestyle{notesep={: }} % 注释分隔符：(Smith, 2020: p.10)

\setcitestyle{authoryear} % Smith (2020) - 默认
\usepackage{algorithm}
\usepackage{algorithmic}

\usepackage{newfloat}
\usepackage{listings}
\DeclareCaptionStyle{ruled}{labelfont=normalfont,labelsep=colon,strut=off} % DO NOT CHANGE THIS
\floatstyle{ruled}
\newfloat{listing}{tb}{lst}{}
\floatname{listing}{Listing}
\newcommand{\revisedhl}[1]

\title{Concise and Sufficient Sub-Sentence Citations for Retrieval-Augmented Generation}
\author{
    Guo Chen\textsuperscript{\rm 1,2},
    Qiuyuan Li\textsuperscript{\rm 1,2},
    Qiuxian Li\textsuperscript{\rm 1,2},
    Hongliang Dai\textsuperscript{\rm 1,2}\thanks{Corresponding author},
    Xiang Chen\textsuperscript{\rm 3},
    Piji Li\textsuperscript{\rm 1,2}
}
\affiliations{
    \textsuperscript{\rm 1}College of Artificial Intelligence,
    Nanjing University of Aeronautics and Astronautics, Nanjing, China\\
    \textsuperscript{\rm 2}The Key Laboratory of Brain-Machine Intelligence Technology, 
    Ministry of Education, Nanjing, China.\\
    \textsuperscript{\rm 3}MIIT Key Laboratory of Pattern Analysis and Machine Intelligence, 
    College of Computer Science and Technology, \\
    Nanjing University of Aeronautics and Astronautics, Nanjing, China\\
    \{162110125, lqy123, 162350227, hongldai,xiang\_chen, pjli\}@nuaa.edu.cn
}

\usepackage{bibentry}
\setcitestyle{aysep={,}} % 作者和年份用逗号分隔：(Smith, 2020)
\usepackage[authoryear]{natbib}  % 关键！启用作者-年份格式
\setcitestyle{round}             % 确保使用圆括号
\usepackage{natbib}
\let\oldthebibliography\thebibliography
\renewcommand{\thebibliography}[1]{%
  \oldthebibliography{#1}%
  \setlength{\itemsep}{0.65ex} % 条目间距
  \setlength{\parsep}{-0.5ex} % 减少条目内段落间距
  \setlength{\itemindent}{-\bibhang}%
  \setlength{\leftmargin}{\bibhang}%
}

\begin{document}

\maketitle

\begin{abstract}
In retrieval-augmented generation (RAG) question answering systems, generating citations for large language model (LLM) outputs enhances verifiability and helps users identify potential hallucinations. However, we observe two problems in the citations produced by existing attribution methods. First, the citations are typically provided at the sentence or even paragraph level. Long sentences or paragraphs may include a substantial amount of irrelevant content. Second, sentence-level citations may omit information that is essential for verifying the output, forcing users to read the surrounding context. In this paper, we propose generating sub-sentence citations that are both concise and sufficient, thereby reducing the effort required by users to confirm the correctness of the generated output. To this end, we first develop annotation guidelines for such citations and construct a corresponding dataset. Then, we propose an attribution framework for generating citations that adhere to our standards. This framework leverages LLMs to automatically generate fine-tuning data for our task and employs a credit model to filter out low-quality examples. Our experiments on the constructed dataset demonstrate that the propose approach can generate high-quality and more readable citations.
\end{abstract}

\section{Introduction}

Despite the rapid advancements and widespread adoption of large language models (LLMs) in recent years, hallucination continues to be a major challenge \citep{azamfirei2023large,ye2023cognitive}. Although various techniques have been proposed for detecting and mitigating hallucinations, fully eliminating them from LLM generated content remains unachievable. As a result, enabling users to efficiently verify the correctness of model outputs has become a critical concern in the deployment of LLM-based applications.

In retrieval augmented generation (RAG)  \citep{lewis2020retrieval}, which supplements LLMs with relevant external context to improve response accuracy, the retrieved content naturally serves as a source for verifying truthfulness \cite{gao2023enabling,shen2025citelab}. However, since the retrieved context can be lengthy, it is often impractical for users to read through all of it.And to some extent, essential verification information may be overlooked. Therefore, recent studies have investigated methods for highlighting or attaching the key portions of the context that support the generated output, thereby reducing the user's verification burden.

One strategy for extracting such key content is to adopt perturbation or gradient-based attribution methods \cite{sudhi2024rag} These methods are able to highlight the key words or phrases in the context to ``explain’’ the model’s output. However, because the highlighted elements are often scattered, the resulting explanations can be difficult for users to interpret efficiently. An alternative approach is to generate citations for the model’s responses directly using the LLM. Existing studies \cite{gao2023enabling,zhang2024longcite} adopting this approach typically produce citations at the paragraph or sentence level. We observe two main limitations with this strategy. First, a cited paragraph or sentence may contain substantial irrelevant information, increasing the user's reading time. Second, the generated citations may omit some important content that are necessary for verification, particularly information about the subject of a cited sentence. This problem forces users to examine the surrounding context of the cited element for clarification.

% \begin{figure}[ht]
%   \begin{minipage}[t]{\columnwidth}
%     \centering
%     \includegraphics[width=\linewidth]{case731.png}
%     \caption{Comparison of annotation strategies}
%     \label{fig:fig1}
%   \end{minipage}
%   \hfill
%   \begin{minipage}[t]{\columnwidth}
%   \end{minipage}
% \end{figure}

\begin{figure}[ht]
    \centering
    \includegraphics[width=\linewidth]{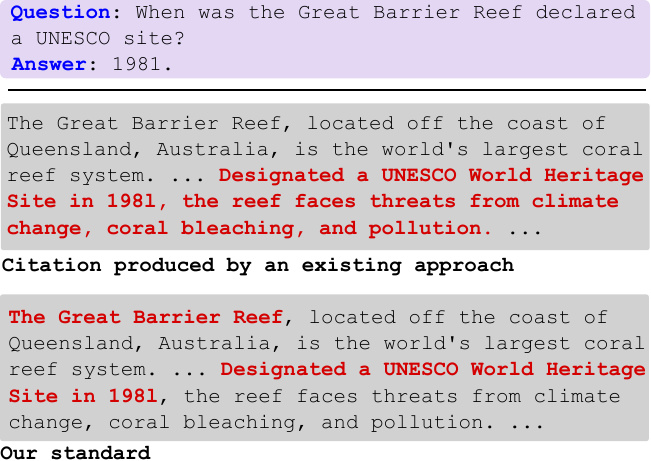}
    \caption{Comparison of cited content produced by an existing approach and our sub-sentence citation standard. The cited content is highlighted in red. Some unimportant content are omitted with ``...''.}
    \label{fig:fig1}
  \hfill
\end{figure}

Therefore, in this paper, we aim to generate sub-sentence level citations in RAG-based systems such that the cited content is more readable, concise, and at the same time sufficient for verifying the LLM responses. To this end, we first we first develop the annotation guidelines for such citations and construct a corresponding dataset. In application, the system may highlight the cited content within the retrieved context, enabling users to verify responses at a glance and significantly improving verification efficiency.

Figure \ref{fig:fig1} shows an example where we compare our citation standard with the result of the existing sentence-level citation generation approach LongCite \cite{zhang2024longcite}. The existing approach cites the full sentence that contains the answer. However, more than half of the content in the sentence is irrelevant. Moreover, it is unclear which reef this sentence is about. Meanwhile, our standard is to cite only the relevant content within the sentence, and also include the clause that contains the name of the referred reef.In this way, the content we annotate can obtain the necessary information for verifying the answer.

% As shown in Figure \ref{fig:fig1}, when the user's question is ``What is the name of the large city in Saudi Arabia in terms of size?'', previous works such as unsupervised method RAG-Ex and supervised method Longcite trained models will use the sentence ``And it is the large city in Saudi Arabia'' from the article as a citation, but it is difficult to clarify the exact referent of the pronoun ``it'' within the sentence due to the lack of explicit supervision signals;  If only the sentence is highlighted as a citation, it is highly likely to have significant negative effects on credibility.Although Longcite and other models can locate text intervals containing answer fragments, they often neglect explicit modeling of referential resolution due to their sentence level granularity annotation. When the answer sentence uses pronouns such as ``it/they/this'' to refer back to the preceding entity, the model's output citation lacks a clear subject, resulting in users still needing to backtrack the entire text to confirm the referent, weakening the self consistency and readability of the citation.

Given the high standard of our annotation guidelines, manual annotation is costly and time-consuming. To address this, we design a data augmentation approach that expands a small manually annotated training set with automatically generated examples. Specifically, we first use LLMs to generate large-scale training data, then filter out low-quality examples using a fine-tuned quality estimation (credit) model. Finally, we fine-tune LLMs to generate citations that meet our standards by training them on both the high-quality machine-generated data and the human-annotated dataset.

We conduct experiments on the dataset we constructed and evaluate our approach using a set of manually annotated examples. The results demonstrate the effectiveness of our method in generating accurate and concise sub-sentence-level citations.

Our contributions are summarized as follows.

\begin{itemize}
    \item We investigate  the generation of concise and sufficient sub-sentence-level citations for LLM-based RAG systems, with an emphasis on alignment with conventional reading patterns. We propose a set of annotation principles that reflect our citation standards and construct a manually annotated dataset accordingly.
    \item We design an approach for generating such citations that requires only a small number of manually annotated training examples.
    \item We conduct experiments with the dataset we construct to verify the effectiveness of our approach. 
\end{itemize}

\section{Related Work} \label{Related Work}

The research on methods for generating attribution texts has gradually become a topic in recent years due to the hallucinations generated by LLMs.
Self-citation is a recently proposed method by \citet{gao2023enabling}, which utilizes the ability of recent LLMs to follow instructions in natural language \citep{raffel2020exploring,chung2024scaling,brown2020language}, thereby avoiding the need for external validators and guiding generic LLMs to generate inline citations with a small number of samples. Self cited answers are generally more relevant to the content provided by the source, but can still contain unsupported statements and inaccurate citations \citep{liu2023evaluating}.  
RAG-Ex \citep{sudhi2024rag} is a model independent and language independent explanatory framework that explores the impact mechanism of context on generated results through diverse input perturbation strategies. Its innovation lies in approximately reconstructing the decision-making process of LLM through systematic perturbation strategies and response comparisons, without relying on internal parameters or structure of the model. However, due to the lack of supervision of display signals, it is susceptible to the influence of contextual document content, resulting in low interpretability of this method. 
The interpretability techniques in Post hoc interpretation methods  \citep{anand2022explainable,kenny2021explaining} can be roughly divided into gradient based methods  \citep{bastings2020elephant,samek2019towards} and perturbation based methods  \citep{bhattacharya2022applied,zafar2021deterministic}, which prolong user waiting time due to the complexity of pipelines. 
The MIRAGE framework \citep{huq2020adversarial} utilizes comparative analysis to predict distribution shifts in the presence or absence of context, identifies sensitive generated lexical elements using KL divergence, and establishes causal relationships between lexical elements and documents using gradient based attribution techniques. However, this method has excessive computational complexity and low robustness. 

LONGCITE \citep{zhang2024longcite} has similarities with our work, proposed a collaborative generation paradigm of ``coarse to fine'' to meet the fine-grained citation requirements in long context scenarios. Due to the fact that the evaluation system is based on sentence granularity, this method may generate overly complex citations in some cases, requiring a significant amount of time to understand the content of the citations. 

Notably, the work contemporaneous with ours by \citet{hirsch2025laquer} also conducts sub-sentence level citations. However, they focus more on how to allow users to actively highlight any fragment (such as a word or phrase) in the generated text, and then provide supportive source texts. They do not study how high quality sub-sentence level cited contents can be obtained.

\section{Data Annotation Principles}
This section introduces the annotation principles used to create a dataset that aligns with our standards for citation generation.

\begin{figure}[ht]
    \centering
    \includegraphics[width=\linewidth]{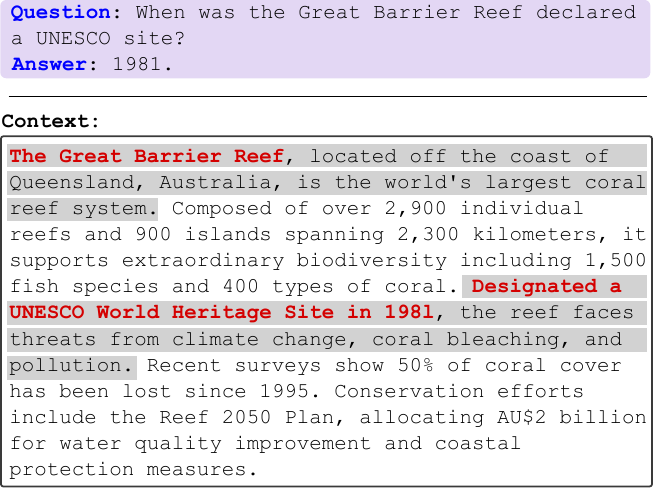}
    \caption{Sub-sentence-level citations adhering to our annotation principles.}
    \label{fig:left}
  \hfill
\end{figure}

Since long texts generated by LLMs can be processed sentence by sentence or fact by fact \cite{min2023factscore,tang2024minicheck}, we only focus on generating citations for simple question answer pairs. Figure \ref{fig:left} offers a more detailed illustration of the earlier example. First, unlike attribution methods that highlight only isolated words (e.g.,  only highlighting ``1981'' in the example), we require that the cited content should be natural, coherent, and easily understandable. To achieve this, we select text fragments that align with conventional reading patterns to provide a smoother reading experience. 

Second, the cited content should be concise, but at the same time sufficient for verifying the answer. In the example of Figure \ref{fig:left}, the sentence containing the answer is ``Designated a UNESCO world Heritage Site in 1981, ... and pollution.'' But the portion beginning with ``the reef faces'' is irrelevant and is therefore omitted from the citation.
Moreover, although this sentence contains the answer, it does not specifically mention which reef is being referred to. Thus, we include the nearest preceding clause that mentions the subject by name, eliminating the need for users to search through earlier sentences for clarification.

% \begin{figure}[ht]
%   \begin{minipage}[t]{\columnwidth}
%     \centering
%     \includegraphics[width=\linewidth]{figure1.png}
%     \caption{Conventional Sentence-granularity Citations and Ours Fixed-granularity Human-comprehensible Attribution}
%     \label{fig:left}
%   \end{minipage}
%   \hfill
%   \begin{minipage}[t]{\columnwidth}
%   \end{minipage}
% \end{figure}

During the data annotation process, we further distinguish between three types of instances.

\begin{itemize}
\item Type-1: The cited content aligns with the conventional sentence-level standard, where a single sentence from the document is highlighted and presented to the user as the citation.
% the content of the citation is consistent with the traditional sentence granularity level method. Highlight a single sentence in the document as a citation provided to the user.
\item Type-2: The cited content would contain irrelevant portions if the instance is annotated according to the conventional sentence-level standard. Such cases commonly arise when the retrieved context originates from formal sources, such as government documents or professional articles.
% it indicates that annotating citations according to traditional sentence granularity levels may appear too redundant, which usually occurs in government documents, professional terminology descriptions, and other fields. So when annotating, only the consecutive tokens in the sentence need to be annotated as citations for the problem. If two disjointed parts of a single sentence appear as citations, it is the third situation.
\item Type-3: Multiple segments of the retrieved context must be cited. This typically occurs in two scenarios: 1) The sentence that supports the answer does not explicitly mention the name of the subject, such as the example in Figure \ref{fig:left}. 2) The answer requires multi-hop inference, necessitating the citation of multiple, dispersed facts from the context to support verification.
% in addition to the two disjointed contents in the single sentence mentioned above, there are also some multi hop problems and problems with contextual interference. If the question provided by the user contains a subject, and words such as ``They'' and ``He'' that refer to the subject with contextual meaning appear during annotation, it is necessary to use LLM's understanding ability to find the corresponding position of the subject in the context, while highlighting the content and location of the subject, as well as the content and location of the answer.
%\revisedhl{(please check the correctness)}
\end{itemize}

To build a dataset that covers the above scenarios, we leverage three existing publicly available question answering datasets including XOR-AttriQA \citep{asai2020xor}, XQUAD \citep{rajpurkar2016squad}, and HotpotQA \citep{yang2018hotpotqa} as the source corpus pool.
Among them, XOR-AttriQA provides attribution annotations required for open-domain QA. But its annotated citations are mainly in sentence-level. After manual secondary refinement, fine-grained fragments directly corresponding to the answers can be obtained. XQUAD is a reading comprehension benchmark with paragraph-level question answering instances. HotpotQA is known for multi-hop reasoning, which is well-suited for verifying the citation localization ability of models in scenarios of long-range dependency and referential resolution. The three datasets complement one another in terms of discourse structure, content diversity, and the depth of reasoning required.

% To build a cross granularity citation dataset that covers multilingual and multi hop inference scenarios, this paper selected XOR-AttriQA \citep{asai2020xor}, XQUAD \citep{rajpurkar2016squad}, and HotpotQA \citep{yang2018hotpotqa} from publicly available high-quality question answering resources as the original corpus pool. 
% Among them, XOR-AttriQA provides attribution annotations required for cross-lingual open-domain QA. Its original corpus is mainly sentence-level citations, and after manual secondary refinement, fine-grained fragments directly corresponding to the answers can be obtained. As a cross-lingual reading comprehension benchmark, XQUAD covers paragraph-level QA in over ten languages and provides rich multilingual materials for extracting and aligning phrase level references; HotpotQA is known for its multi-hop reasoning and complex document relationships, which is precisely used to verify the citation localization ability of the model in scenarios of long-range dependency and referential resolution. The three types of datasets complement each other in terms of language, discourse structure, and inference depth. After unified screening, alignment, and re-annotation, they together constitute a diverse and consistent experimental resource, laying a solid foundation for the evaluation of fine-grained attribution frameworks.

% \subsection{Data Generation}
\section{Methodology}

\begin{figure}[ht]
  \centering
  \includegraphics[width=\linewidth]{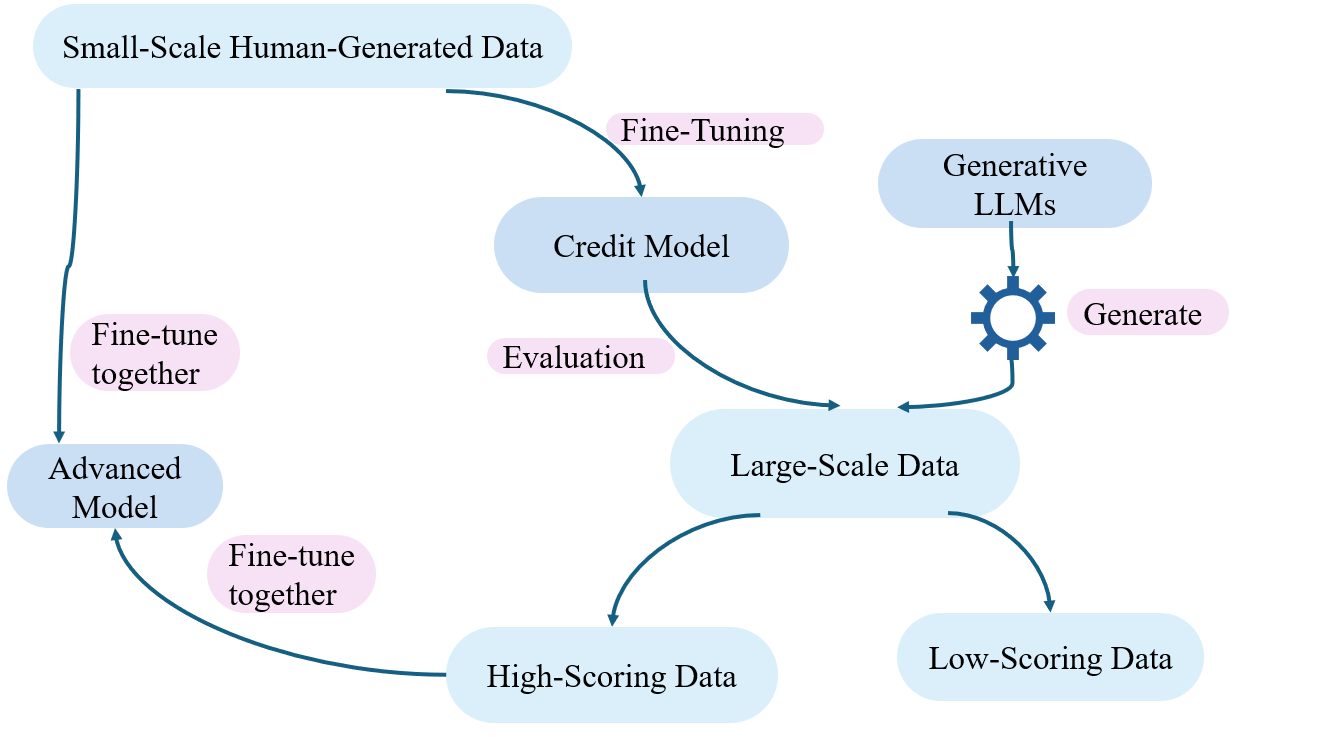}
  \caption{Overall procedure of our approach.}
  \label{fig:wide}
\end{figure}

% \revisedhl{(overall procedure of the approach)}
% In this section, we will introduce our methodology. 
Since manually annotating examples that meets our citation standards is costly and time-intensive, we introduce an approach to produce training data automatically by leveraging LLMs. The overall procedure is illustrated in Figure \ref{fig:wide}.

First, we use a high-quality dataset manually annotated based on our annotation principles to fine tune LLM, which serves as a fine tuned quality estimation (credit) model. Then, using prompt word templates, the text generation class model generates a large amount of training data based on the dataset examples we provide, and then uses the fine tuned credit model to filter out low-quality examples. Subsequently, the system calls open-source generation models to generate candidate question answer citation pairs in bulk under a unified hierarchical constraint prompt framework. These candidate samples are first fed into the credit model for confidence scoring. High scoring examples are directly included in the next round of fine-tuning corpus, while low scoring examples are downgraded to coarse-grained. Through this cycle of ``small models learning from large models and large models refining small models'', the dataset improves synchronously in both scale and quality, significantly expanding the training volume and domain coverage, while ensuring the consistency of fine-grained standards. Finally, we fine tuned LLM by training it on high-quality machine generated data and human annotated datasets to generate references that meet our standards.

During the dataset expansion phase, we integrate multiple open-source LLMs (such as DeepSeek-R1 and GPT-3.5) through APIs to enhance data diversity and expand scale. The model is guided to generate training data with fine-grained citations through structured prompt engineering. The prompt we adopt includes task instructions, input and output of the training set, strictly following the requirements to generate data entries. The input includes complete paragraphs and clear questions, and the output field must be an exact reference fragment that directly supports the answer in the original text. The design goal of the prompt template is to ensure that the generated citations meet two core requirements: (1) accurate support for answer content; (2) Maintain fluency for human readability. 
% To further achieve domain adaptation, the system supports users to upload target domain literature as contextual prior knowledge, driving large models to generate data samples that conform to specific domain characteristics, thereby constructing a domain customized training resource library.

%\revisedhl{(details about the number of training examples and the models used should be moved to the Experiments section)}
%Before starting large-scale data generation, we take 500 cross-granularity citation samples that have been manually annotated as the starting point, and uses LoRA fine-tuning strategy to train a lightweight ``credit model'' on the Llama-3.1-8B base. This model has the comprehensive ability to judge answer accuracy, citation granularity, and readability with only a few updated parameters. Subsequently, the system calls open-source generation models to generate candidate question answer citation pairs in bulk under a unified hierarchical constraint prompt framework. These candidate samples are first fed into the credit model for confidence scoring. High scoring examples are directly included in the next round of fine-tuning corpus, while low scoring examples are downgraded to coarse-grained. Through this cycle of ``small models learning from large models and large models refining small models'', the dataset improves synchronously in both scale and quality, significantly expanding the training volume and domain coverage, while ensuring the consistency of fine-grained standards.

\section{Experiments}
In order to verify the effectiveness of the fine-grained attribution framework based on large models proposed in this article in practical applications, this section presents experiments to validate the effectiveness of our annotation method after fine-tuning the model. Before starting large-scale data generation, we take 800 cross-granularity citation samples that have been manually annotated as the starting point,the ratio of this dataset under the three types we set is 15\%, 30\%, and 50\%. and uses LoRA fine-tuning strategy to train a lightweight ``credit model'' on the Llama-3.1-8B base. This model has the comprehensive ability to judge answer accuracy, citation granularity, and readability with only a few updated parameters.
\begin{table}[h]
\centering
\setlength{\tabcolsep}{3pt}      % 列间距再紧一点
\footnotesize                     % 比 \small 更小
\begin{tabular}{lcc}
\toprule
Type & \# Examples & Ratio\\
\midrule
Type-1 & 120 & 15\% \\
Type-2 & 280 & 35\% \\
Type-3 & 400 & 50\% \\
\midrule
\bfseries Total & \bfseries 800 & \bfseries 100\%\\
\bottomrule
\end{tabular}
\caption{Statistics of the dataset.}
\label{tab:dataset-split}
\end{table}
The comprehensive performance of the system evaluation model in three dimensions: fine simplification of citation granularity, semantic consistency, and human readability. The Longcite method is a model obtained by fine-tuning a large number of sentence granularity level QA problems. The RAG-Ex method is a model independent perturbation based approach.The experiment used LLaMa3.1-8b and Qwen2.5-7b as baseline models, Longcite model and unsupervised RAG-Ex method as baseline methods for the case study, and a two-stage fine-tuning strategy and data co evolution mechanism to iteratively generate a series of models that can generate concise and sufficient citations. In the first and second stages of fine-tuning, the ratio of the training set to the test set is 80\% and 20\%, respectivelyThe training framework is based on DeepSpeed and LoRA technology stacks, ensuring efficient parameter updates under limited video memory constraints.

\subsection{Evaluation Metrics}
In the data evaluation stage, this study uses a three-layer complementary indicator system to quantify the alignment between model output and manual annotation: surface vocabulary matching is undertaken by F1 score, distribution-level semantic consistency is characterized by cosine similarity, and the subjective dimension of generated quality is assigned to GPT-4o for overall scoring. The F1 score is used as a comprehensive indicator to measure the degree of lexical overlap between predicted text and reference text, and its core idea stems from the harmonic average of precision and recall in information retrieval and classification tasks. Specifically for the citation generation task involved in this study, the calculation of F1 first relies on lexical alignment of the text sequence: the system splits the original string into token sequences through regular expressions and eliminates case differences in lowercase space to establish comparable units. Subsequently, the word element sets of the predicted sequence and the reference sequence were separately constructed as unordered sets to remove the influence of word frequency and retain only the morphological information. Based on this set, the system calculates the intersection size of the two as the true positive (TP). Precision is defined as the proportion of correct word elements in the predicted text, i.e. the ratio of TP to the predicted set cardinality; The recall rate is defined as the proportion of successfully covered lexical elements in the reference text, that is, the ratio of TP to the reference set cardinality. The F1 \eqref{eqn-1} score combines the two through harmonic averaging, and its closed-form expression is as follows:

\begin{equation}\label{eqn-1}
\mathrm{F1}=2\cdot\frac{\mathrm{P}\cdot\mathrm{R}}{\mathrm{P}+\mathrm{R}},
\end{equation}

\begin{equation}\label{eqn-1}
\mathrm{Precision}=\frac{|\,P\cap R\,|}{|\,P\,|},\quad
\mathrm{Recall}=\frac{|\,P\cap R\,|}{|\,R\,|}.
\end{equation}
The symbols P and R represent the sets of predicted and referenced lexical elements, respectively. The numerical range is between 0 and 1, and the closer it is to 1, the more complete the overlap between the predicted text and the reference text in terms of word form. This indicator is used in this experiment to quantify the surface-level consistency between the model output and manually annotated citations, especially for evaluating whether the answer covers key entities and key vocabulary, without involving deep semantic matching. Parallel to this, cosine similarity introduces weight information based on the bag of words model: the system first maps the predicted text and the reference text to the same high-dimensional word frequency vector space, with corresponding vectors denoted as p and r. Then, the cosine value of the angle between the two is calculated as following:
\begin{equation}\label{eqn-1}
\cos\theta
=\frac{\mathbf{p}\cdot\mathbf{r}}{\|\mathbf{p}\|\,\|\mathbf{r}\|}
=\frac{\sum_{i=1}^{n}p_i r_i}{\sqrt{\sum_{i=1}^{n}p_i^{2}}\sqrt{\sum_{i=1}^{n}r_i^{2}}}.
\end{equation}
Among them, $p^{i}$ and $r^{i}$ are the normalized word frequencies of the i-th morpheme. This value measures the directional consistency of word frequency distribution by the ratio of dot product to modulus length, and is robust to changes in word order and syntax, thus capturing semantic approximations that may be overlooked by the surface F1 \eqref{eqn-1}. Finally, the automatic evaluation of GPT-4o takes structured prompts as input and comprehensively considers the three dimensions of accuracy, conciseness, and readability. Each citation is given a comprehensive score in the range of 0-1, thereby transforming human preferences into reproducible quantitative signals. The combination of the three not only covers multiple levels of granularity from word form to semantics to human perception, but also avoids evaluation bias that may be caused by a single indicator, providing a robust empirical basis for the effectiveness of fine-grained attribution frameworks.

\subsection{Fine-tuned Model Performance Validation}
To evaluate the effectiveness of the fine-grained attribution framework, this study conducted fine-tuning experiments on the LLaMa3.1-8b and Qwen2.5-7b models based on LoRA low rank adaptation technology. The experiment adopts a parameter configuration with Rank=8 and Scaling Factor=32. Under the premise of freezing the original model parameters, only the incremental matrix is trained to adapt to the fine-grained citation generation task. The training data includes 500 manually annotated samples and 1000 level extended samples filtered by credit models, with fine-grained samples accounting for a stable proportion of over 80\%. The gradient accumulation steps are set to 8, the batch size is 2, the optimizer uses AdamW (learning rate 5e-5), and the maximum sequence length is truncated to 1024 words.
\begin{itemize}
\item Word level F1 score: used to calculate the overlap rate between predicted citations and manually annotated vocabulary, measuring surface matching accuracy;
\item Semantic cosine similarity: Capturing deep semantic consistency through embedding vector distribution alignment;
\item GPT-4o Comprehensive Quality Score: Based on a structured prompt template, weighted fusion accuracy, conciseness, and readability scores in three dimensions,the scoring process formula is as follows:
\begin{equation}\label{eqn-2}
(Q,A,C_{pred})\rightarrow GPT\text{-4o}
\end{equation}
Formula \eqref{eqn-2} is the input for the evaluation process, where Q represents the problem, A represents the answer, and C represents the citation. This formula combines questions, answers, and citations into a structured input that is provided to the GPT\text{-4o} model for scoring.
\begin{equation}\label{eqn:1}
\tilde{S}_{i} = \frac{S_{i} - 1}{4}
\end{equation}
\begin{equation}\label{eqn:quality}
S_{\text{quality}} = 
\lambda_1 \tilde{S}_{\text{acc}} +
\lambda_2 \tilde{S}_{\text{conc}} +
\lambda_3 \tilde{S}_{\text{read}}
\end{equation}
\begin{equation}\label{eqn:weight}
\lambda_1 + \lambda_2 + \lambda_3 = 1, \quad
\lambda_i \ge 0,\; i \in \{1,2,3\}
\end{equation}
The scoring dimensions include three points, namely accuracy $S_{acc}$ : the strength of causal relationship between citations and answers; $S_{conc}$: Information density and redundancy; $S_{read}$: The fluency of language expression, where formula \eqref{eqn:1} is the output obtained by scoring, which is a comprehensive evaluation of the quality of citations and reflects their performance in three dimensions: accuracy, conciseness, and readability. Formula \eqref{eqn:quality} is a constraint condition of the scoring formula, used to ensure the rationality of the scoring. This is to ensure the scientificity and effectiveness of the scoring process. This constraint ensures that the score S remains within the range of 0 to 1. The closer the score is to 1, the better the quality of the citation. If the score is closer to 0, the worse the quality of the citation.
\end{itemize}
\begin{table}[h]
\centering % 表格居中
\begin{tabular}{cccc} % 定义表格列数为4
\toprule % 表格顶部横线
\textbf{Model} & \textbf{F1} & \textbf{CS} & \textbf{GPT-4o} \\ % 表头
\midrule % 表格中部横线
Qwen2.5-7B & 0.4616 & 0.5542 & 0.4839 \\
Subcite-Qwen2.5-7B & \textbf{0.7319} & \textbf{0.7977} & \textbf{0.7624} \\ \hline
LlaMa3.1-8B & 0.3976 & 0.4692 & 0.4358 \\
LongCite-llama3.1-8B & 0.5328 & 0.6021 & 0.5637 \\
Subcite-llama3.1-8B & \textbf{0.6547} & \textbf{0.7336} & \textbf{0.6953} \\
\bottomrule % 表格底部横线
\end{tabular}
\caption{Comparison of model performance} % 表格标题
\label{tab4.1:Performance} 
\end{table}

Table \ref{tab4.1:Performance} presents the evaluation experimental results of LlaMa3.1-8B and Qwen2.5-7B models before and after fine-tuning. After fine-tuning, the models showed certain improvements in various indicators. It should be noted that after fine-tuning, the F1 index of Qwen2.5-7B increased to 0.7319, which is a significant improvement compared to the original base model. This confirms that the fine-grained support capability has been significantly improved. This improvement is mainly due to the framework's ability to accurately identify the minimum sufficient evidence set, while the learning mechanism of semantic associations between answers and citations also plays a role. In the cross model comparison process, Qwen2.5-7B showed a smaller performance gap compared to LlaMa3.1-8B with more parameters after the same training process, and the framework had excellent model generalization ability. This proves that this mechanism has good adaptability to small and medium-sized models, providing a feasible solution for deployment in resource limited scenarios.The comparative evaluation results between our model and the Longcite method model are shown in Table \ref{tab4.1:Performance}. It can be seen that in terms of annotation principles in this article, our evaluation effect is relatively good.

\subsection{Case Study on Fine-grained Attribution}
\begin{table*}[t]
\centering
\small
\renewcommand{\arraystretch}{1.2}
\begin{tabular}{@{}p{2.6cm} p{6.5cm} p{6.5cm}@{}}
\toprule
\multirow{2}{*}{\textbf{Methods}} &
\textbf{Q1: ``What is the capital of the state of Assam?''} &
\textbf{Q2: ``How many colors are used in the national flag of Southern Mexico?''}\\
\cmidrule(lr){2-2} \cmidrule(lr){3-3}
 & \multicolumn{1}{c}{\textbf{Generated Answer}} & \multicolumn{1}{c}{\textbf{Generated Answer}} \\
\midrule
RAG-Ex &
Dispur is the capital of the state of Assam in India. Dispur, a locality of Guwahati, became the capital of Assam in 1973. This was after Shillong, the erstwhile capital,became the capital of the state of Meghalava that was carved out of Assam,. Dispur is the seat of Government of Assam. The Assam Secretariat building is located in dispur along with the Assam Assembly House,MLA Hostels and the State Emergency Operations Centre, The Assam Trunk road and the G s road passes through Dispur. To the south ofDispur is the theologically important site of Basistha. … &
The President of Guatemala since then. \textcolor{red}{The flag is divided in four parts, red, yellow, white and black, each colour representing Xinca people, Garifuna people, Maya peoples and Ladino people, respectively.}These colours are also part of the "Q'anil", a Maya symbol in which each color represents a point of the compass, an element of nature and a part of the human being. "Q'anil" means "seed" in Maya script, and is also used for one of the 20 days of the Maya calendar.  … \\[2mm]
\midrule
Longcite Model &
\textcolor{red}{Dispur is the capital of the state of Assam in India. Dispur, a locality of Guwahati, became the capital of Assam in 1973.} This was after Shillong, the erstwhile capital,became the capital of the state of Meghalava that was carved out of Assam,. Dispur is the seat of Government of Assam. The Assam Secretariat building is located in dispur along with the Assam Assembly House,MLA Hostels and the State Emergency Operations Centre, The Assam Trunk road and the G s road passes through Dispur. To the south ofDispur is the theologically important site of Basistha. … &
The President of Guatemala since then. \textcolor{red}{The flag is divided in four parts, red, yellow, white and black, each colour representing Xinca people, Garifuna people, Maya peoples and Ladino people, respectively.}These colours are also part of the "Q'anil", a Maya symbol in which each color represents a point of the compass, an element of nature and a part of the human being. "Q'anil" means "seed" in Maya script, and is also used for one of the 20 days of the Maya calendar.  … \\[2mm]
\midrule
Ours &
\textcolor{red}{Dispur is the capital of the state of Assam in India.} Dispur, a locality of Guwahati, became the capital of Assam in 1973. This was after Shillong, the erstwhile capital,became the capital of the state of Meghalava that was carved out of Assam,. Dispur is the seat of Government of Assam. The Assam Secretariat building is located in dispur along with the Assam Assembly House,MLA Hostels and the State Emergency Operations Centre, The Assam Trunk road and the G s road passes through Dispur. To the south ofDispur is the theologically important site of Basistha. … &
The President of Guatemala since then. \textcolor{red}{The flag is divided in four parts, red, yellow, white and black}, each colour representing Xinca people, Garifuna people, Maya peoples and Ladino people, respectively.These colours are also part of the "Q'anil", a Maya symbol in which each color represents a point of the compass, an element of nature and a part of the human being. "Q'anil" means "seed" in Maya script, and is also used for one of the 20 days of the Maya calendar.  … \\[2mm]
\bottomrule
\end{tabular}
\caption{Comparison of citation results generated by different methods on two queries}
\label{tab:two_examples}
\end{table*}

When the user's question is `` What is the capital of the state of Assam? '', the citation results obtained by the three methods are shown in Table \ref{tab:two_examples}. In an open domain Q\&A scenario, when there are two independent and informative sentences in the context that can independently support the same answer, the traditional RAG Ex single sentence perturbation strategy may not provide effective confidence due to ``single point failure": no matter which sentence is removed, the remaining information is still sufficient to answer the question, resulting in insignificant perturbation differences and inability to determine the true supporting sentence. However, the supervision method Longcite model highlights all sentences that can provide answers as citations, which is a redundant behavior. When users verify the accuracy of answers, they can obtain answers with only one sentence, without the need to spend time looking at the second citation. Therefore, in this case, the citations obtained by this method appear very complicated.Similarly, when the user's question is ``How many colors are used in the national flag of Southern Mexico?'', other methods may also exhibit redundant answers.The answer obtained through our method is a precise statement that can fully answer the user's question without any unnecessary information, which provides convenience for users to verify their answers.

\subsection{Ablation Study on Data Scalability}
\begin{table}[h]
\centering
\begin{tabular}{lcc}
\toprule
\textbf{Model} & \textbf{Expaned Data Size} & \textbf{F1} \\
\midrule
\multirow{3}{*}{Subcite2.5-7B} 
 & 500 & 0.7319 \\
 & 700 & 0.7387 \\
 & 1000 & 0.7653 \\
\bottomrule
\end{tabular}
\caption{Model Performance with Varying Sample Sizes}
\label{tab43:sample_performance}
\end{table}
In this section, we conducted an experiment on the impact of adding an expanded dataset on the experimental results. Table \ref{tab43:sample_performance}. shows the ablation experiment results of the training sample size for Qwen2.5-7B model. Further analysis of the impact of training data size reveals the effectiveness of the data iteration mechanism through experiments on the scalability of training data size. The F1 training result obtained when only using the initial manually annotated 500 data points is 0.7319. After adding a large number of fine-grained samples generated by the large model, the F1 obtained is 0.7387, which is an improvement compared to before. This also proves that the model data co evolution mechanism is feasible. When the training samples increased from 500 to 1000, the F1 index maintained a stable upward trend, verifying the data self enhancement ability of the framework.
The F1 score reached 0.7653 at 1000 samples, which is an improvement from the benchmark of 500 samples.

The reason why we did not conduct experiments on a large number of sentence granularity level datasets organized by adding other existing open source datasets is mainly due to the fact that when the model comes into contact with sentence granularity level training samples, it will learn two different attribution patterns: one is based on fine-grained attribution, and the other is based on coarse-grained attribution, which will affect the updating of model parameters.

\section{Conclusion}
%\revisedhl{(conclusion is too long)}
This article proposes an improvement approach for annotating training data, focusing on the issues of coarse citations and time-consuming verification in attribution fine-tuning in the Retrieval-Augmented Generation Question Answering system. The approach involves using a small amount of manual annotation to initiate the process, followed by self expansion and self inspection using open-source large models.  

The data expansion process of the framework can increase the size of the dataset, but optimization is needed to ensure data quality. The integration of the dataset is not precise enough, and adding new data directly can easily lead to redundancy and inconsistency. The current framework mainly deals with text data. When facing scenes containing multimodal information such as images and audio, it is difficult to effectively integrate different modal data to improve attribution performance. In the future, cross modal alignment methods based on geometric deep learning can be explored to study how to map different modal information to a unified space and achieve more comprehensive attribution analysis. In terms of processing extremely long texts, the current framework is not very efficient. When faced with lengthy documents, the model may experience performance degradation during retrieval and citation generation. There are also areas for improvement in the evaluation system. Currently, the proposed indicators mainly focus on English QA issues, and their adaptability needs to be improved in cross language scenarios. Subsequently, efforts can be made to establish a multilingual attribution benchmark test set that includes multiple language types and text forms, while simultaneously developing a dynamic verification protocol based on causal reasoning to enhance the comprehensiveness and accuracy of the evaluation.
\bibliography{AnonymousSubmission/LaTeX/aaai2026}
\end{document}